\PassOptionsToPackage{table,xcdraw}{xcolor}  
\documentclass[sigconf]{acmart}
\AtBeginDocument{%
  }

\usepackage{multirow}
\usepackage{graphicx}
\usepackage{booktabs}
\usepackage{makecell}
\usepackage{enumitem}
\usepackage{hyperref}
\usepackage[table,xcdraw]{xcolor}  
\setcopyright{acmlicensed}
\copyrightyear{2025}
\acmYear{2025}
\setcopyright{acmlicensed}
\acmConference[MM '25] {Proceedings of the 33rd ACM International Conference on Multimedia}{October 27--31, 2025}{Dublin, Ireland.}
\acmBooktitle{Proceedings of the 33rd ACM International Conference on Multimedia (MM '25), October 27--31, 2025, Dublin, Ireland}
\acmISBN{979-8-4007-2035-2/2025/10}
\acmDOI{10.1145/3746027.3755340}

\acmSubmissionID{3303}

\begin{document}

\title{Grounding Emotion Recognition with Visual Prototypes: VEGA - Revisiting CLIP in MERC}

\author{Guanyu Hu}
\orcid{0009-0006-1911-6283}
\affiliation{%
  \institution{Xi'an Jiaotong University}
  \city{Xi'an}
  \country{China}
}
\affiliation{%
  \institution{Queen Mary University of London}
  \city{London}
  \country{UK}
}
\email{g.hu@qmul.ac.uk}

\author{Dimitrios Kollias}
\authornote{Corresponding author.}
\orcid{0009-0006-1911-6283}
\affiliation{%
  \department{Center for Multimodal AI}
  \department{Digital Environment Research Institute}
  \institution{Queen Mary University of London}
  \city{London}
  \country{UK}
  }
\email{d.kollias@qmul.ac.uk}

\author{Xinyu Yang}
\affiliation{%
  \institution{Xi'an Jiaotong University}
  \city{Xi'an}
  \country{China}
}
\email{yxyphd@mail.xjtu.edu.cn}


\begin{abstract}
    Multimodal Emotion Recognition in Conversations remains a challenging task due to the complex interplay of textual, acoustic and visual signals. While recent models have improved performance via advanced fusion strategies, they often lack psychologically meaningful priors to guide multimodal alignment. In this paper, we revisit the use of CLIP and propose a novel Visual Emotion Guided Anchoring (VEGA) mechanism that introduces class-level visual semantics into the fusion and classification process. Distinct from prior work that primarily utilizes CLIP’s textual encoder, our approach leverages its image encoder to construct emotion-specific visual anchors based on facial exemplars. These anchors guide unimodal and multimodal features toward a perceptually grounded and psychologically aligned representation space, drawing inspiration from cognitive theories (prototypical emotion categories and multisensory integration). A stochastic anchor sampling strategy further enhances robustness by balancing semantic stability and intra-class diversity. Integrated into a dual-branch architecture with self-distillation, our VEGA-augmented model achieves sota performance on IEMOCAP and MELD. Code is available at: \href{https://github.com/dkollias/VEGA}{https://github.com/dkollias/VEGA}.

\end{abstract}




\maketitle

\section{Introduction}
Emotion Recognition in Conversations (ERC) presents unique challenges due to the dynamic and spontaneous nature of dialogues, where individuals express a wide range of emotions that evolve over time \cite{poria2019emotion,zheng2023facial,hu2024robust,wei2023multi}. While early ERC approaches predominantly relied on textual data \cite{li2021semi}, this modality alone often fails to capture the subtlety and variability of emotional expression \cite{hazarika2018icon}. As a result, Multimodal ERC (MERC), which incorporates audio and visual modalities alongside text, has gained growing interest for its potential to model richer emotional cues \cite{hu2022mm,yang2023self,wei2022audio}.

To integrate information from multiple modalities, MERC methods commonly adopt fusion strategies such as feature concatenation \cite{tu2022exploration,hazarika2018icon}, attention mechanisms \cite{yang2023self,shi2023multiemo} and heterogeneous graph-based approaches \cite{chen2023multivariate,hu2021mmgcn,hu2022mm}. Recent state-of-the-art models, such as \cite{zhang2023cross}, have advanced this space by hierarchically integrating modality-specific features and optimizing for emotion classification. While these methods have led to strong performance on benchmark datasets, they remain fundamentally data-driven and lack structured, semantically aligned priors to guide the fusion process. This limits their robustness and generalization, especially when handling ambiguous, noisy, or weakly correlated multimodal inputs, which are common in real-world conversational settings.

In this paper, we introduce a novel approach that enhances the semantic alignment and robustness of multimodal fusion by integrating CLIP-guided visual anchors, which serve as emotion-specific prototypes encoded from real images. For each of the emotional classes, we select a small set of representative facial images and encode them using CLIP’s image encoder. The mean embedding per class is used as a semantic anchor, which guides the alignment of all three modalities (visual, acoustic, textual) within a shared emotion-centric representation space. To enhance generalization and mitigate overfitting to fixed anchors, we introduce a stochastic sampling strategy that alternates between using class-wise center anchors and randomly selected instances during training. This approach balances semantic stability with intra-class variability, effectively serving as semantic augmentation. The sampled anchors are applied to both unimodal and fused representations, promoting discriminative, robust, and semantically coherent feature learning.

Importantly, while most prior approaches that leverage CLIP focus on its textual modality to align features, typically mapping audio or visual inputs to text-derived embeddings (e.g., prompts such as ``this is a sad face''), we deliberately reverse this trend.
These prior methods use text prompts because CLIP’s textual encoder offers a powerful, zero-shot compatible representation space and is easier to manipulate through language. However, textual descriptions of emotions are often ambiguous, abstract, and culturally variable, whereas emotional expressions in human faces are more concrete, universal, and visually grounded~\cite{fehr1984concept,shaver1987emotion,hu2025rethinking,hu2024bridging,wei2024learning,kollias2024mma}. By anchoring fusion around CLIP’s visual embeddings, we embrace a more perceptually grounded and psychologically aligned representation of emotion, closer to how humans perceive affective states.

This design draws on rich insights from cognitive psychology. The use of CLIP-based visual anchors reflects the prototypical emotion theory \cite{ekman1992argument,ekman1994basic}, which posits that emotional categories are organized around prototypical exemplars, such as a canonical ``angry'' or ``happy'' face. The use of CLIP-derived anchors mimics this structure by introducing emotion prototypes derived from real-world facial imagery. 
Moreover, our approach is aligned with the theory of multisensory integration in emotion perception \cite{degelder2000perception}, which suggests that the human brain aligns information from multiple channels using known emotional schemas. By aligning each modality with shared visual anchors, our method mimics this natural mechanism of emotional cognition. 
Additionally, affective priming~\cite{fazio2001automatic} and embodied cognition theories~\cite{barsalou2008grounded} provide psychological justification: humans often interpret ambiguous emotional signals by referencing mental imagery or contextually induced affective states. Our visual anchors serve a similar role in guiding the model’s internal representation toward consistent emotional interpretations.

By bridging advances in vision-language pretraining with psychologically grounded theories of emotion, our approach offers a new paradigm for MERC that is both more robust and more human-aligned. In summary, our contributions are threefold:
\begin{itemize}[leftmargin=10pt]

    \item \textbf{Architecture-Agnostic Visual Emotion Guided Anchoring (VEGA):}  
    We propose the VEGA mechanism with a Stochastic Anchor Sampling Strategy to construct emotion-specific visual anchors that align all modalities (text, audio, visual) within a shared, semantically structured space. VEGA is fully modular and can be seamlessly integrated into existing MERC pipelines without architectural changes.
    \item \textbf{Reversing the CLIP Paradigm:}  
    In contrast to prior work that primarily aligns visual features to CLIP’s textual embeddings through prompt-based supervision, our approach reverses this paradigm by anchoring fusion in CLIP’s visual embedding space. To the best of our knowledge, this is the \textbf{\textit{first}} visually driven use of CLIP in MERC, opening a new and unexplored direction.

    \item \textbf{Psychologically Grounded Design:}  
    Our framework is inspired by cognitive and affective psychological theories, such as prototypical emotion categories and multisensory integration, providing semantically meaningful guidance for emotion modeling.

    \item \textbf{State-of-the-Art Performance:}  
    Our approach achieves state-of-the-art results on two widely used benchmarks, IEMOCAP and MELD, demonstrating its effectiveness in both fine-grained and diverse conversational emotion recognition settings.

\end{itemize}

\section{Related Work}

\textit{Multimodal Emotion Recognition in Conversations (MERC).}  
While traditional ERC has focused on textual input~\cite{poria2019emotion,pereira2022deep}, its limited emotional expressiveness has driven the development of MERC, which integrates text, audio, and visual modalities~\cite{li2021semi,zhang-li-2023,kollias20246th,kollias20247th,kollias2025advancements,zadeh2018memory,hu2021dialoguegcn,yoon2018multimodal}. Early fusion approaches~\cite{li2021semi} simply concatenate features, but often struggle with modality imbalance and noise. More advanced strategies leverage attention~\cite{yoon2018multimodal,zhao2023swrr}, memory~\cite{zadeh2018memory}, context-aware modeling~\cite{majumder2019dialoguernn,hu2021dialoguegcn}, and semantic refinement~\cite{zhang2023multimodal,zhang-li-2023} to better capture inter-modal and temporal dependencies. However, most existing models remain data-driven, lacking structured semantic priors or emotion-level prototypes to guide multimodal alignment. 

\textit{CLIP for Emotion and Multimodal Representation.}  
CLIP~\cite{radford2021clip} has demonstrated strong generalization in vision-language tasks via large-scale contrastive pretraining. In emotion recognition, recent works~\cite{emovclip2023,zhao2023prompting,qi2024multimodal} adapt CLIP's textual encoder with emotion prompts for zero-shot classification, yet focus on text-based alignment. These approaches underutilize CLIP’s visual encoder and overlook the universality of visual affective cues. Notably, CLIP has not been explored in the context of MERC, leaving a gap in leveraging perceptual visual semantics for multimodal emotional grounding.

\section{Methodology}

\begin{figure*}[htbp]
  \centering
  \includegraphics[page=1,width=\linewidth]{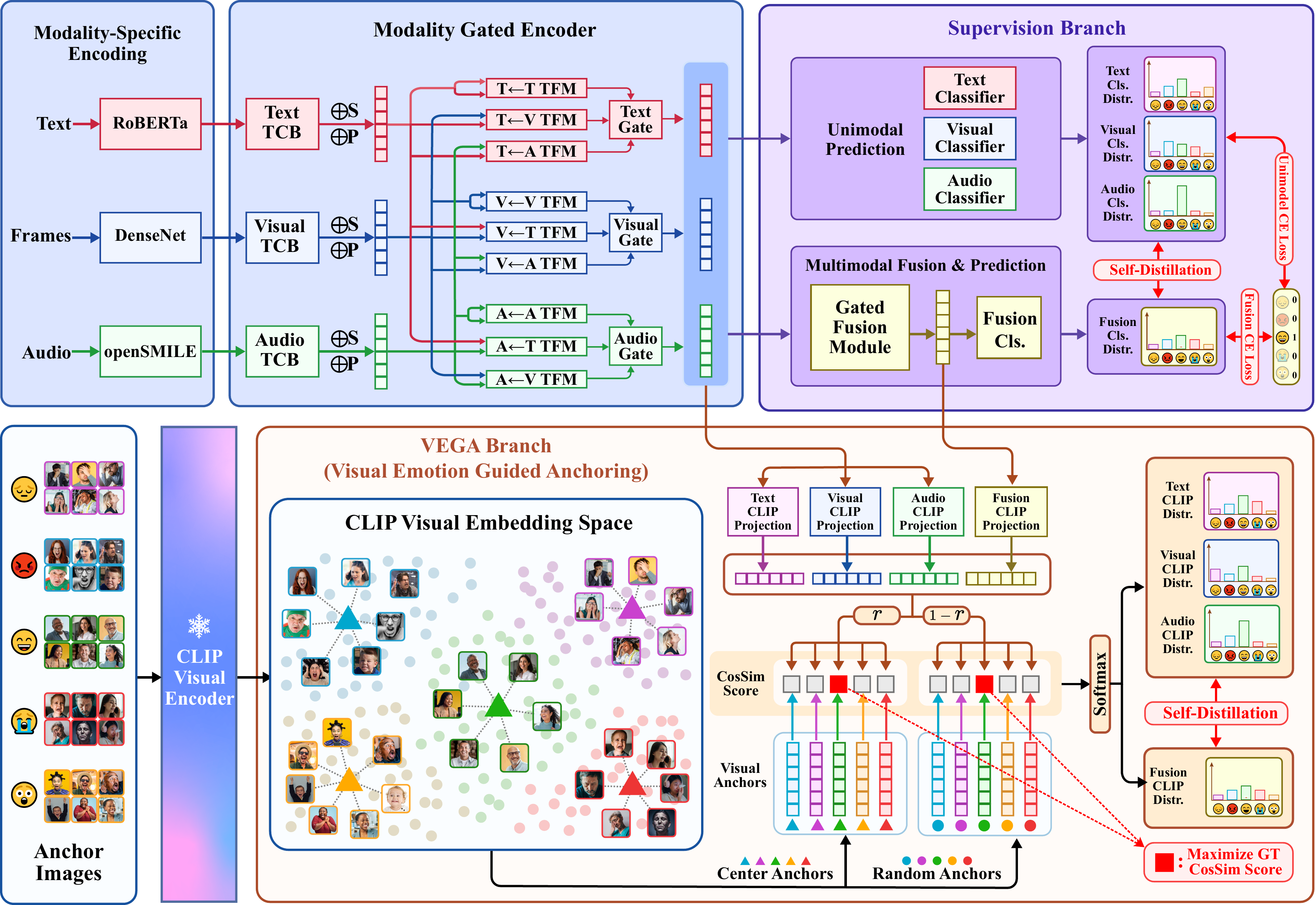}

\caption{Overview of SDT-VEGA, comprising of: (i) Modality-Specific Encoding; (ii) Modality Gated Encoder; Dual-Branch Hierarchical Anchoring and Supervision, including (iii) Supervision Branch and (iv) VEGA Branch. The lower part illustrates the VEGA module, which takes both unimodal and fused features as inputs and performs Visual Emotion Guided Anchoring with CLIP-based emotion anchors. Here, TCB denotes  Temporal Convolution Block, TFM refers to modality-specific Transformer modules described in Section~\ref{sec:modality} and T/V/A represent  text, visual and audio modalities, respectively,  defined in Eq.~\ref{eq:attention}.}

\end{figure*}

In the following, we present MERC, which builds upon the Self-Distillation Transformer \cite{ma2023transformer} and comprises the following components:
\textbf{(i) Modality-Specific Encoding:}  
Independently encodes unimodal features from each input modality (text, audio, and visual).
\textbf{(ii) Modality Encoder:}  
Captures both intra-modal and inter-modal contextual dependencies using parallel Transformer-based encoders.
\textbf{(iii) Dual-Branch Hierarchical Anchoring and Fusion:}  
Progressively constructs semantically enriched unimodal and multimodal representations via two parallel branches.  
The \textit{supervised branch} performs gated feature fusion followed by label-supervised classification, while the \textit{VEGA branch} applies CLIP-based Visual Emotion Guided Anchoring (VEGA) to align features with external visual anchors derived from CLIP image embeddings.
\textbf{(iv) Classification and Self-Distillation:}  
Both the \textit{supervised branch} and the \textit{VEGA branch} independently perform classification and incorporate self-distillation to enforce cross-modal consistency.

Let a conversation be represented as a sequence of \( N \) utterances, \( C = \{ \mathbf{u}_1, \mathbf{u}_2, \ldots, \mathbf{u}_N \} \).  
Each utterance \( \mathbf{u}_t \) is defined as a tuple of modality-specific inputs:  
$
u_t = \left( \mathbf{x}_t^{(T)}, \mathbf{x}_t^{(A)}, \mathbf{x}_t^{(V)} \right),
$
where \( \mathbf{x}_t^{(T)} \), \( \mathbf{x}_t^{(A)} \), and \( \mathbf{x}_t^{(V)} \) represent the raw textual, acoustic, and visual signals of the \( t \)-th utterance, respectively.
The objective is to predict the emotion label \( y_t \in \mathcal{Y} \) corresponding to each utterance \( \mathbf{u}_t \).

\subsection{Modality-Specific Encoding}

Each input modality is first passed through a dedicated encoder to extract utterance-level unimodal features:
\begin{equation}
\mathbf{h}_t^{(m)} = \mathcal{F}_{\text{enc}}^{(m)}(\mathbf{x}_t^{(m)}), \quad m \in \{T, A, V\}
\end{equation}
where \( \mathcal{F}_{\text{enc}}^{T} \), \( \mathcal{F}_{\text{enc}}^{A} \), and \( \mathcal{F}_{\text{enc}}^{V} \) refer to RoBERTa, OpenSMILE, and DenseNet-based encoders, respectively. The output \( \mathbf{h}_t^{(m)} \in \mathbb{R}^{d_m} \) is the initial unimodal representation of utterance \( \mathbf{u}_t \) for modality \( m \).

\subsection{Modality Gated Encoder and Fusion}
\label{sec:modality}

To capture local temporal patterns, we apply \( \mathcal{F}_{\text{conv}}(\cdot) \), a Temporal Convolution Block (TCB), to extracted features:
\begin{equation}
\tilde{\mathbf{h}}_t^{(m)} = \mathcal{F}_{\text{conv}} \left( \mathbf{h}_{t-k}^{(m)}, \ldots, \mathbf{h}_{t+k}^{(m)} \right),
\end{equation}
where: \( k \) is the temporal window size and $\tilde{\mathbf{h}}_t^{(m)} \in \mathbb{R}^{d} $.

Next, we inject structural information into each modality stream by adding:  
(i) positional embeddings \( \mathbf{p}_t \), which encode the position of each utterance within the conversation; and  
(ii) speaker embeddings \( \mathbf{s}_t \), which indicate the speaker identity and help capture speaker-specific emotional dynamics:
\begin{equation}
\hat{\mathbf{h}}_t^{(m)} = \tilde{\mathbf{h}}_t^{(m)} + \mathbf{p}_t + \mathbf{s}_t.
\end{equation}

These enhanced modality-specific representations are further processed by a set of parallel Transformer (TFM) modules to model both intra-modal and inter-modal contextual dependencies.  
For each modality \( m \in \{T, A, V\} \), we employ three contextual Transformer modules:  
(i) an \textbf{intra-modal Transformer} \( \mathcal{T}_{\text{intra}}^{(m \leftarrow m)} \), which captures both temporal and contextual dynamics within modality \( m \);  
(ii) an \textbf{inter-modal Transformer} \( \mathcal{T}_{\text{inter}}^{(m \leftarrow m_1)} \), which models modality-specific cross-modal interactions from modality \( m_1 \) to \( m \); and  
(iii) another \textbf{inter-modal Transformer} \( \mathcal{T}_{\text{inter}}^{(m \leftarrow m_2)} \), which encodes complementary information from modality \( m_2 \) to \( m \),  
where \( \{ m_1, m_2 \} = \{T, A, V\} \setminus \{m\} \).  
Each module outputs a conversation-level contextual representation for modality \( m \) as follows:
\begin{align}
\tilde{\mathbf{z}}_{\text{intra}}^{(m)} &= \mathcal{T}_{\text{intra}}^{(m \leftarrow m)}\left( \hat{\mathbf{h}}^{(m)};\ \hat{\mathbf{h}}^{(m)} \right), \\
\tilde{\mathbf{z}}_{\text{inter1}}^{(m)} &= \mathcal{T}_{\text{inter}}^{(m \leftarrow m_1)}\left( \hat{\mathbf{h}}^{(m)};\ \hat{\mathbf{h}}^{(m_1)} \right), \\
\tilde{\mathbf{z}}_{\text{inter2}}^{(m)} &= \mathcal{T}_{\text{inter}}^{(m \leftarrow m_2)}\left( \hat{\mathbf{h}}^{(m)};\ \hat{\mathbf{h}}^{(m_2)} \right),
\label{eq:attention}
\end{align}
where \( \left\{ \tilde{\mathbf{z}}_{\text{intra}}^{(m)},\ \tilde{\mathbf{z}}_{\text{inter1}}^{(m)},\ \tilde{\mathbf{z}}_{\text{inter2}}^{(m)} \right\} \subset \mathbb{R}^{N \times d} \).  
This enables each modality stream to integrate multi-perspective contextual information.


To further enhance representation quality, we apply a modality-specific gating mechanism that adaptively filters each context stream based on its semantic relevance. For each \( \tilde{\mathbf{z}}^{(m)}_{*, t} \in \tilde{\mathbf{z}}^{(m)}_{*} \), a learnable linear transformation followed by a sigmoid activation is applied:
\begin{equation}
\mathbf{z}_{*, t}^{(m)} = \sigma \left( \mathbf{W}_{*}^{(m)} \tilde{\mathbf{z}}_{*, t}^{(m)} \right) \odot \tilde{\mathbf{z}}_{*, t}^{(m)}, 
\end{equation}
where, $* \in \{ \text{intra}, \text{inter1}, \text{inter2} \}$ and \( \odot \) denotes element-wise multiplication and \( \sigma(\cdot) \) is the sigmoid function.
Finally, the gated outputs are concatenated and passed through a modality-specific linear layer \( \mathcal{F}_{\text{cat}}^{(m)}(\cdot) \) to obtain the final unified unimodal representation:
\begin{equation}
\mathbf{z}^{(m)} = \mathcal{F}_{\text{cat}}^{(m)}\left( \left[ \mathbf{z}_{\text{intra}}^{(m)} \,\|\, \mathbf{z}_{\text{inter1}}^{(m)} \,\|\, \mathbf{z}_{\text{inter2}}^{(m)} \right] \right), \quad \mathbf{z}^{(m)} \in \mathbb{R}^{N \times d} 
\end{equation}


\subsection{Dual Branch for Supervision and Anchoring}

After obtaining the final unimodal-level representations, architecture bifurcates into two semantically distinct branches:
\textbf{(i) Supervision Branch}: This branch conducts label-supervised classification at both unimodal and multimodal levels. It adaptively integrates the modality-specific representations to produce a final prediction aligned with the ground-truth emotion labels.
\textbf{(ii) VEGA Branch}: The CLIP-based Visual Emotion Guided Anchoring (VEGA) branch projects both modality-specific and fused representations into the CLIP visual embedding space, aligning them with emotion-specific visual anchors. This branch provides high-level semantic supervision by encouraging the learned features to semantically align with human-interpretable emotion prototypes.

The introduction of two parallel branches offers several advantages over a single-branch design:   
\textit{1) Decoupled Objectives:} Visual anchoring and supervised classification are optimized independently, mitigating gradient interference between semantic alignment and task-specific learning.
\textit{2) Modularity and Scalability:} The VEGA branch can be fine-tuned or extended in isolation, without requiring retraining of the fusion branch, thus enabling flexible experimentation and deployment.
\textit{3) Complementary Supervision:} While the fusion branch focuses on capturing signal quality and cross-modal interactions, the VEGA branch grounds the learning process in semantically rich visual priors, enhancing robustness under noisy or ambiguous input conditions.

\subsubsection{\textbf{\underline{Supervision Branch}}}

This branch introduces both unimodal and fusion-level supervision to enhance discriminability of modality-specific representations and the robustness of multimodal fusion.

\vspace{0.2cm}
\noindent
\textbf{Unimodal Prediction.}  
Each modality-specific representation \( \mathbf{z}^{(m)} \) is passed through a linear classifier \( \mathcal{C}_{\text{uni}}^{(m)}(\cdot) \) to generate prediction:
\begin{align}
    \hat{\mathbf{y}}^{(m)} = \text{Softmax}\left( \mathcal{C}_{\text{uni}}^{(m)}(\mathbf{z}^{(m)}) \right) , \quad \hat{y}^{(m)} = \arg\max_{c \in \mathcal{Y}} \left[ \hat{\mathbf{y}}^{(m)} \right]_c.
\end{align}

\vspace{0.2cm}
\noindent
\textbf{Multimodal Gated Fusion \& Prediction.}  
To integrate information across modalities, a soft gating mechanism \( \mathcal{G}(\cdot) \) is used to assign importance weights to each modality. The fused representation \( \mathbf{f} \in \mathbb{R}^d \) is computed as:
\begin{align}
    \mathbf{f} = \sum_{m} \frac{\exp\left( \mathcal{G}(\mathbf{z}^{(m)}) \right)}{\sum_{m'} \exp\left( \mathcal{G}(\mathbf{z}^{(m')}) \right)} \cdot \mathbf{z}^{(m)}.
\end{align}
The fusion classifier \( \mathcal{C}_{\text{fuse}}(\cdot) \) is used to produce the prediction:
\begin{align}
    \hat{\mathbf{y}} = \text{Softmax}\left( \mathcal{C}_{\text{fuse}}(\mathbf{f}) \right), \quad 
    \hat{y} = \arg\max_{c \in \mathcal{Y}} \left[ \hat{\mathbf{y}} \right]_c.
\end{align}

\noindent
\textbf{Supervision Objectives.}  
This branch is optimized using a combination of hard-label and soft-label (self-distillation) objectives.

\textit{Hard-Label Supervision.}  
Cross-entropy loss is computed for both unimodal and fused predictions with respect to the ground-truth label \( y \in \mathcal{Y} \):
\begin{align}
    \mathcal{L}_{\text{cls}}^{(m)} &= -\log \left( \left[ \hat{\mathbf{y}}^{(m)} \right]_y \right), \\
    \mathcal{L}_{\text{cls}}^{\text{fuse}} &= -\log \left( \left[ \hat{\mathbf{y}} \right]_y \right).
\end{align}

\textit{Self-Distillation (Soft-Label Supervision).}  
To encourage semantic consistency between unimodal and multimodal representations, we adopt a KL-based distillation loss that aligns unimodal predictions with the fused prediction:
\begin{align}
    \mathcal{L}_{\text{dist}}^{(m)} = \text{KL} \left( \hat{\mathbf{y}} \,\|\, \hat{\mathbf{y}}^{(m)} \right).
\end{align}

\noindent
\textit{Total Supervision Loss.}  
The total loss for the Supervision Branch is a weighted combination of classification and distillation objectives:
\begin{align}
    \mathcal{L}_{\text{sup}} = \lambda_{\text{cls}}^{\text{fuse}} \cdot\mathcal{L}_{\text{cls}}^{\text{fuse}} + \sum_{m} \left( \lambda_{\text{cls}}^{(m)} \cdot\mathcal{L}_{\text{cls}}^{(m)} + \lambda_{\text{dist}} \cdot \mathcal{L}_{\text{dist}}^{(m)} \right),
\end{align}
where \( \lambda_{\text{dist}} \) is a tunable hyperparameter controlling the strength of the distillation regularization.

\subsubsection{\textbf{\underline{VEGA Branch}}}

So far, our framework operates in a fully data-driven manner and lacks explicit semantic priors to support emotion understanding. To address this limitation, we introduce a \textit{CLIP-based Visual Emotion Guided Anchoring} mechanism, which injects semantic structure by leveraging visual emotion anchors derived from CLIP-encoded facial images. These anchors serve as global semantic references for aligning learned representations.

VEGA is applied at two levels of representation:  
\textbf{(i) Unimodal Anchoring:} VEGA guides the semantic projection of each modality-specific representation, enhancing the emotional discriminability of unimodal features.  
\textbf{(ii) Multimodal Anchoring:} After fusion, the same anchoring mechanism is applied to refine the fused representation, aligning it with CLIP-derived emotion anchors in a shared semantic space.  
By enforcing semantic alignment with shared emotion anchors at both unimodal and multimodal level, VEGA promotes feature semantic consistency, strengthens cross-modal grounding, and improves generalization.

\vspace{0.2cm}
\noindent
\textbf{Anchor Construction:}
For each emotion class \( c \in \mathcal{Y} \), we collect a small set of representative facial images \( \{ I_c^{(1)}, I_c^{(2)}, \ldots, I_c^{(n)} \} \). Each image is encoded using the frozen CLIP image encoder \( \mathcal{E}_{\text{CLIP}}^{\text{img}}(\cdot) \):
\begin{align}
    \mathbf{v}_c^{(i)} = \mathcal{E}_{\text{CLIP}}^{\text{img}}(I_c^{(i)}), \quad i = 1, \ldots, n
    \label{eq:anchor}
\end{align}
The class-wise \textit{Center Anchor} is derived by averaging the $n$ embedding vectors associated with each emotion class:
\begin{align}
    \bar{\mathbf{v}}_c = \frac{1}{n} \sum_{i=1}^{n} \mathbf{v}_c^{(i)}
    \label{eq:center}
\end{align}
Each \( \bar{\mathbf{v}}_c \in \mathbb{R}^{d_{\text{anc}}} \) serves as a visual emotion anchor, capturing high-level, human-aligned semantic structure. These anchors are used to guide the alignment of modality-specific and fused representations within a shared emotion-centric representation space.

\vspace{0.2cm}
\noindent
\textbf{Stochastic Anchor Sampling Strategy.}
To improve the stability and generalization of visual emotion guided anchoring, we propose a simple yet effective stochastic anchor sampling strategy. During training, we alternate between two types of anchors for each emotion class: the \textit{Center Anchor} (the mean embedding computed as in Eq.\eqref{eq:center}) and the \textit{Random Anchor} (a randomly selected instance embedding from the same class as defined in Eq.\eqref{eq:anchor}). This stochasticity encourages the model to align features not only with stable semantic prototypes but also with diverse intra-class variations, thereby enhancing robustness and flexibility. 

This strategy offers two key benefits: \textbf{(i) Stability} — the center anchor serves as a canonical prototype, providing a stable and generalizable semantic reference for alignment, thereby promoting semantic consistency and reducing sensitivity;  \textbf{(ii) Variability} — the use of randomly selected anchors introduces semantic perturbations, encouraging the model to learn robust associations between diverse intra-class visual representations and their emotion labels. This stochastic alternation between stable and augmentative anchors acts as a form of semantic augmentation, mitigating overfitting to specific anchor instances and enhancing semantic diversity for more flexible and generalizable feature learning.

Formally, at each training iteration, for a given emotion class \( c \), the selected anchor \( \mathbf{a}_c \in \mathbb{R}^{d_{\text{anc}}} \) is defined as:
{\small\begin{align}
    \mathbf{a}_c =
    \begin{cases}
        \bar{\mathbf{v}}_c & \text{if } r = 1 \quad \text{(center anchor)} \\
        \mathbf{v}_c^{(j)}, \quad j \sim \text{Uniform}(1, n) & \text{if } r = 0 \quad \text{(random anchor)}
    \end{cases}
\end{align}}
To express the strategy more compactly, we define:
{\small\begin{align}
    \mathbf{a}_c = r \cdot \left( \bar{\mathbf{v}}_c  \right) + (1 - r) \cdot \mathbf{v}_c^{(j)},
\end{align}} \noindent
where \( r \sim \text{Bernoulli}(q) \) is a binary variable determining the sampling mode. The hyperparameter \( q \in (0, 1) \) controls the probability of selecting the center anchor (e.g., \( q = 0.5 \) gives equal probability to both types). \( j \in \{1, \ldots, n\} \) denotes a randomly sampled index.

\vspace{0.2cm}
\noindent
\textbf{Visual Emotion Guided Anchoring.}  
We apply the VEGA mechanism at both the unimodal and multimodal levels to inject CLIP-derived visual emotion anchor priors into the representation space.

\textit{Unimodal Anchoring.}  
Each modality-specific representation \( \mathbf{z}^{(m)} \) is first projected into the CLIP-aligned embedding space using a learnable projection module \( \mathcal{P}_{\text{uni}}(\cdot) \):
\begin{align}
    \tilde{\mathbf{z}}^{(m)} = \mathcal{P}_{\text{uni}}(\mathbf{z}^{(m)}), \quad \tilde{\mathbf{z}}^{(m)} \in \mathbb{R}^{d_{\text{anc}}}.
\end{align}
The cosine similarity between the projected representation and each emotion anchor \( \mathbf{a}_c \) is computed as:
\begin{align}
    s_c^{(m)} = \cos(\tilde{\mathbf{z}}^{(m)}, \mathbf{a}_c) = \frac{ \tilde{\mathbf{z}}^{(m)} \cdot \mathbf{a}_c }{ \| \tilde{\mathbf{z}}^{(m)} \| \cdot \| \mathbf{a}_c \| }, \quad \forall c \in \mathcal{Y}.
\end{align}
These similarity scores are then normalized via softmax to obtain the anchor-based emotion prediction distribution:
\begin{align}
    \hat{y}_{\text{anc},c}^{(m)} = \frac{\exp(s_c^{(m)})}{\sum_{c' \in \mathcal{Y}} \exp(s_{c'}^{(m)})}, \quad \forall c \in \mathcal{Y},
\end{align}

\textit{Multimodal Anchoring.}  
The fused representation \( \mathbf{f} \) is projected using a separate fusion-specific module \( \mathcal{P}_{\text{fuse}}(\cdot) \):
\begin{align}
    \tilde{\mathbf{f}} = \mathcal{P}_{\text{fuse}}(\mathbf{f}), \quad \tilde{\mathbf{f}} \in \mathbb{R}^{d_{\text{anc}}}.
\end{align}
Its similarity with each anchor is computed by:
\begin{align}
    s_c^{\text{fuse}} = \cos(\tilde{\mathbf{f}}, \mathbf{a}_c) = \frac{ \tilde{\mathbf{f}} \cdot \mathbf{a}_c }{ \| \tilde{\mathbf{f}} \| \cdot \| \mathbf{a}_c \| }, \quad \forall c \in \mathcal{Y}.
\end{align}
The normalized fusion-level anchor prediction is given by:
\begin{align}
    \hat{y}_{\text{anc},c}^{\text{fuse}} = \frac{\exp(s_c^{\text{fuse}})}{\sum_{c' \in \mathcal{Y}} \exp(s_{c'}^{\text{fuse}})}, \quad \forall c \in \mathcal{Y},
\end{align}

\noindent
\textbf{VEGA Objectives.}  
Similar to the Supervision Branch, VEGA incorporates both hard-label objectives and soft-label (self-distillation) objectives at the unimodal and fusion levels.

\textit{Modality-Specific Anchoring Loss.}  
For each modality \( m \), we compute the anchor-based classification loss using the cross-entropy between the predicted distribution \( \hat{\mathbf{y}}_{\text{anc}}^{(m)} \) and the ground-truth label \( y \in \mathcal{Y} \):
\begin{align}
    \mathcal{L}_{\text{anc}}^{(m)} = -\log \left( \hat{y}_{\text{anc},y}^{(m)} \right),
\end{align}
where \( \hat{y}_{\text{anc},y}^{(m)} \) denotes the predicted probability of the correct class.

\textit{Multimodal Anchoring Loss.}  
A similar classification loss is defined for the fused representation:
\begin{align}
    \mathcal{L}_{\text{anc}}^{\text{fuse}} = -\log \left( \hat{y}_{\text{anc},y}^{\text{fuse}} \right).
\end{align}

\textit{Anchoring Self-Distillation Loss.}  
To promote semantic consistency across modalities in the VEGA-aligned space, we introduce a distillation loss where the fusion prediction \( \hat{\mathbf{y}}_{\text{anc}}^{\text{fuse}} \) acts as a teacher to guide each modality-specific anchor prediction:
\begin{align}
    \mathcal{L}_{\text{anc-dist}}^{(m)} = \text{KL} \left( \hat{\mathbf{y}}_{\text{anc}}^{\text{fuse}} \,\|\, \hat{\mathbf{y}}_{\text{anc}}^{(m)} \right).
\end{align}

\noindent
\textit{Total VEGA Loss.}  
The total loss for the VEGA Branch combines the above terms:
\begin{align}
    \mathcal{L}_{\text{vega}} = \lambda_{\text{anc}}^{\text{fuse}} \cdot \mathcal{L}_{\text{anc}}^{\text{fuse}} + \sum_{(m)} \left( \lambda_{\text{anc}}^{m} \cdot \mathcal{L}_{\text{anc}}^{(m)} + \lambda_{\text{anc-dist}} \cdot \mathcal{L}_{\text{anc-dist}}^{(m)} \right).
\end{align}

\subsection{Overall Objective}

The overall training objective combines the Supervision Branch and VEGA Branch losses:
\begin{align}
    \mathcal{L}_{\text{total}} = \mathcal{L}_{\text{sup}} + \mathcal{L}_{\text{vega}},
\end{align}

\section{Experimental Results and Discussion}

\begin{table*}[ht]
\centering
\caption{Performance comparison on IEMOCAP and MELD between our proposed method and sota. Best results are in bold.}
\label{sota}
\resizebox{0.95\textwidth}{!}{
\begin{tabular}{lcccccccccccccccc}
\toprule
\multirow{4}{*}{\textbf{Model}} & \multicolumn{14}{c}{\textbf{IEMOCAP}} & \multicolumn{2}{c}{\textbf{MELD}} \\
\cmidrule(lr){2-15} \cmidrule(lr){16-17}
& \multicolumn{2}{c}{\textbf{Happy}} & \multicolumn{2}{c}{\textbf{Sad}} & \multicolumn{2}{c}{\textbf{Neutral}} & \multicolumn{2}{c}{\textbf{Angry}} & \multicolumn{2}{c}{\textbf{Excited}} & \multicolumn{2}{c}{\textbf{Frustrated}} &\multirow{2}{*}{\textbf{ACC}}   &\multirow{2}{*}{\textbf{F1}} & \multirow{2}{*}{\textbf{ACC}} & \multirow{2}{*}{\textbf{F1}} \\
\cmidrule(lr){2-3}
\cmidrule(lr){4-5}
\cmidrule(lr){6-7}
\cmidrule(lr){8-9}
\cmidrule(lr){10-11}
\cmidrule(lr){12-13}
& \textbf{ACC} & \textbf{F1} & \textbf{ACC} & \textbf{F1} & \textbf{ACC} & \textbf{F1} & \textbf{ACC} & \textbf{F1} & \textbf{ACC} & \textbf{F1} & \textbf{ACC} & \textbf{F1} & & & & \\
\midrule
\textbf{COGMEN}~\cite{joshi2022cogmen}                    & --                   & 51.90                & --                     & 81.70                & --                   & 68.60                & --                     & 66.00                  & --                   & 75.30                & --                    & 58.20                & 68.20                & 67.60             & --                    & -- \\
\textbf{MMGCN}~\cite{hu2021mmgcn}                         & 32.64                & 39.66                & 72.65                  & 76.89                & 65.10                & 62.81                & 73.53                  & \textit{71.43}         & 77.93                & 75.40                & 65.09                 & 63.43                & 66.61                & 66.25             & --                    & 58.65 \\
\textbf{MM-DFN}~\cite{hu2022mm}                           & 44.44                & 44.44                & 77.55                  & 80.00                & 71.35                & 66.99                & \textbf{75.88}         & 70.88                  & 74.25                & 76.42                & 58.27                 & 61.67                & 67.84                & 67.85             & 62.49                 & 59.46 \\
\textbf{DialogueTRM}~\cite{mao2020dialoguetrm}            & 54.19                & 48.70                & 71.04                  & 77.52                & \underline{77.10}    & \textit{74.12}       & 64.74                  & 66.27                  & 63.91                & 70.24                & \underline{70.71}     & 67.23                & 68.92                & 69.23             & 65.66                 & 63.55 \\
\textbf{MMTr}~\cite{zou2022improving}                     & --                   & --                   & --                     & --                   & --                   & --                   & --                     & --                     & --                   & --                   & --                    & --                   & 72.27                & 71.91             & 64.64                 & 64.41 \\
\textbf{GraphSmile}~\cite{li2024tracing}                  & --                   & \textit{63.09}       & --                     & \textit{83.16}       & --                   & 71.07                & --                     & 71.38                  & --                   & 79.66                & --                    & 66.84                & 72.77                & \textit{72.81}    & \underline{67.70}     & \textit{66.71} \\
\textbf{MA-CMU-SGRNet}~\cite{zhang2024multi}              & 52.60                & 57.10                & 78.80                  & 79.90                & 74.30                & 71.00                & \underline{75.20}      & \underline{71.50}      & \underline{80.30}    & 78.40                & 65.10                 & 67.50                & 72.40                & 71.60             & --                    & 62.30  \\
\textbf{GCCL}~\cite{dai2024multi}                         & --                   & 54.05                & --                     & 81.10                & --                   & 70.28                & --                     & 68.21                  & --                   & 72.17                & --                    & 64.00                & 69.87                & 69.29             & 62.82                 & 60.28  \\
\textbf{D$^{2}$GNN}~\cite{dai2024multimodal}              & --                   & 61.11                & --                     & \underline{83.19}    & --                   & 68.22                & --                     & 66.12                  & --                   & 75.22                & --                    & 63.73                & 70.22                & 69.77             & 61.72                 & 59.74  \\
\textbf{GraphCFC}~\cite{li2023graphcfc}                   & --                   & 43.08                & --                     & \textbf{84.99}       & --                   & 64.70                & --                     & 71.35                  & --                   & 78.86                & --                    & 63.70                & 69.13                & 68.91             & 61.42                 & 58.86  \\
\textbf{AGF-IB}~\cite{shou2024adversarial}                & \textit{71.88}       & \textbf{69.96}       & 74.64                  & 81.58                & 67.25                & 63.80                & \textit{73.79}         & 68.37                  & \textbf{82.66}       & 79.15                & 60.45                 & 63.95                & 70.46                & 70.36             & 64.14                 & 64.01  \\
\textbf{AdaIGN}~\cite{tu2024adaptive}                     & --                   & 53.04                & --                     & 81.47                & --                   & 71.26                & --                     & 65.87                  & --                   & 76.34                & --                    & 67.79                & --                   & 70.74             & --                    & \underline{66.79}  \\
\textbf{MGLRA}~\cite{meng2024masked}                      & 62.90                & 63.50                & \underline{81.10}      & 81.50                & 70.90                & 71.50                & 60.20                  & 61.10                  & 74.40                & 76.30                & \textit{69.20}        & 67.80                & 71.30                & 70.10             & 66.40                 & 64.90  \\
\textbf{MKIN-MCL}~\cite{shen2024multimodal}               & --                   & 58.44                & --                     & 78.17                & --                   & 82.85                & --                     & 68.44                  & --                   & 70.73                & --                    & \underline{70.71}    & \textit{72.86}       & 72.59             & 65.67                 & 64.90  \\
\textbf{DQ-Former}~\cite{jing2024dq}                      & --                   & 71.87                & --                     & 61.67                & --                   & 69.61                & --                     & \textbf{77.39}         & --                   & \underline{80.95}    & --                    & 67.71                & 71.68                & 71.76             & 64.88                 & 64.70  \\
\textbf{Frame-SCN}~\cite{shi2025multimodal}               & 48.95                & --                   & \textbf{83.91}         & --                   & 67.90                & --                   & 70.21                  & --                     & 78.50                & --                   & 68.61                 & --                   & 70.98                & 71.00             & 61.92                 & 59.10  \\
\textbf{SDT}~\cite{sdt}                                   & \underline{72.71}    & \textit{66.19}       & 79.51                  & 81.84                & \textit{76.33}       & \underline{74.62}    & 71.88                  & 69.73                  & 76.79                & \textit{80.17}       & 67.14                 & \textit{68.68}       & \underline{73.95}    & \underline{74.08} & \textit{67.55}        & 66.60 \\
\midrule                                                                                                                                                                                                                                                                                                                                                                                                                    
\rowcolor{gray!20}   
\textbf{VEGA (Ours)}                         & \textbf{74.11}       & \underline{68.81}    & \textit{80.09}         & 82.40                & \textbf{78.22}       & \textbf{75.13}       & 66.43                  & 69.81                  & \textit{80.07}       & \textbf{81.10}       & \textbf{72.27}        & \textbf{70.39}       & \textbf{76.02}       & \textbf{75.58}    & \textbf{69.72}        & \textbf{68.54} \\
\bottomrule
\end{tabular}
}
\end{table*}

\begin{table*}[t]
\centering
\caption{
Performance comparison on IEMOCAP across three unimodal baselines in both standard and VEGA-enhanced versions. Reported metrics include class-wise accuracy (ACC), F1-score (F1), overall accuracy (ACC) and weighted-average F1-score (w-F1).
}
\label{baseline}
\resizebox{\textwidth}{!}{
\begin{tabular}{llcccccccccccccc}
\toprule
\multirow{2}{*}{\textbf{Modality}} &\multirow{2}{*}{\textbf{Model Configuration}} & \multicolumn{2}{c}{\textbf{Happy}} & \multicolumn{2}{c}{\textbf{Sad}} & \multicolumn{2}{c}{\textbf{Neutral}} & \multicolumn{2}{c}{\textbf{Angry}} & \multicolumn{2}{c}{\textbf{Excited}} & \multicolumn{2}{c}{\textbf{Frustrated}} & \multirow{2}{*}{\textbf{ACC}} & \multirow{2}{*}{\textbf{w-F1}} \\
\cmidrule(lr){3-4}
\cmidrule(lr){5-6}
\cmidrule(lr){7-8}
\cmidrule(lr){9-10}
\cmidrule(lr){11-12}
\cmidrule(lr){13-14}
 & & \textbf{ACC} & \textbf{F1} & \textbf{ACC} & \textbf{F1} & \textbf{ACC} & \textbf{F1} & \textbf{ACC} & \textbf{F1} & \textbf{ACC} & \textbf{F1} & \textbf{ACC} & \textbf{F1} & & \\
\midrule
\multirow{2}{*}{\textbf{Visual}} 
& DenseNet+Transformer         & 17.36          & 22.52          & 47.35          & 43.77          & 25.00          & 28.96          & \textbf{41.76} & 36.69          & \textbf{74.58} & \textbf{67.27} & \textbf{45.14} & \textbf{44.05} & 43.31          & 42.03 \\
& DenseNet+Transformer+VEGA    & \textbf{40.28} & \textbf{34.22} & \textbf{52.65} & \textbf{45.58} & \textbf{41.41} & \textbf{42.86} & 38.24          & \textbf{41.01} & 57.53          & 59.31          & 38.32          & 41.60          & \textbf{44.92} & \textbf{45.04} \\
\midrule
\multirow{2}{*}{\textbf{Text}} 
& RoBERTa+Transformer          & \textbf{65.28} & 52.22          & \textbf{79.18} & 79.51          & 62.24          & 64.51          & \textbf{65.88} & 64.00          & 70.23          & 72.16          & 58.53          & 61.52          & 66.05          & 66.34 \\
& RoBERTa+Transforme+VEGA     & 63.89          & \textbf{58.04} & 78.37          & \textbf{80.17} & \textbf{64.58} & \textbf{64.75} & 62.94          & \textbf{65.44} & \textbf{75.92} & \textbf{76.05} & \textbf{64.57} & \textbf{64.74} & \textbf{68.52} & \textbf{68.63} \\
\midrule
\multirow{2}{*}{\textbf{Audio}} 
& OpenSmile+Transformer        & \textbf{47.92} & \textbf{39.66} & 61.63          & 66.23          & 54.69          & 59.32          & \textbf{76.47} & 57.91          & 62.21          & \textbf{69.02} & 50.39          & 51.47          & 57.79          & 58.42 \\
& OpenSmile+Transformer+VEGA    & 36.81          & 38.69          & \textbf{69.39} & \textbf{70.98} & \textbf{70.05} & \textbf{65.77} & 70.00          & \textbf{61.18} & \textbf{62.54} & 66.55          & \textbf{50.92} & \textbf{53.59} & \textbf{61.12} & \textbf{60.96} \\
\bottomrule
\end{tabular}}

\end{table*}

\subsection{Datasets and Evaluations}

\textit{Datasets.}
We conduct experiments on  IEMOCAP~\cite{iemocap} and MELD~\cite{meld} datasets.
IEMOCAP consists of dyadic conversations between ten professional actors, comprising a total of 153 dialogues and 7,433 utterances. Each utterance is manually segmented and annotated with categorical emotion labels. The dataset is divided into five sessions, with each session featuring a unique pair of speakers.
MELD contains 1,433 multi-party conversations and 13,708 utterances, each labeled with one of seven emotion categories.

\textit{Evaluation Protocol.}
Following standard practice in previous works~\cite{sdt}, we split the dataset into training, validation and test sets. We report class-wise accuracy (ACC) and F1-score (F1), as well as overall accuracy (ACC) and weighted-average F1 (w-F1) to comprehensively evaluate model performance.

\subsection{Implementation Details}
Our model is implemented in PyTorch and optimized using AdamW. The initial learning rate is $3 \times 10^{-4}$ with a weight decay of $7 \times 10^{-1}$. We use batch size 15, and apply dropout 0.5 in  the main encoder and in classification head. The Transformer module is configured with 8 attention heads and a hidden dimension of 1280. For CLIP-guided anchoring, we extract emotion anchors using 35 reference images per class. The CLIP projection head consists of 2 layers with SiLU activation and a dropout rate of 0.4. During training, we employ stochastic anchor sampling with a probability threshold $q = 0.2$.  The weights in the loss have been set empirically as: $\lambda_{\text{cls}}^{\text{fuse}} = \lambda_{\text{cls}}^{(m)} = 0.5$, $\lambda_{\text{anc}}^{\text{fuse}} = \lambda_{\text{anc}}^{(m)} = \lambda_{\text{anc-dist}} = 0.6$, and $\lambda_{\text{dist}}  = 0.9$. All reported results are averaged over 10 independent runs with different random seeds.

\subsection{Comparison with state-of-the-art (SOTA)}

We compare our method against SOTA on both IEMOCAP and MELD, as shown in Table~\ref{sota}. The evaluated methods span a diverse range of paradigms, including early fusion (e.g., MM-DFN), graph-based reasoning (e.g., D$^{2}$GNN, MMGCN, Frame-SCN, GraphSmile), memory-enhanced dialogue modeling (e.g., MGLRA, COGMEN), attention-based architectures (e.g., MKIN-MCL), and transformer-based models (e.g., DQ-Former, DialogueTRM, SDT). To ensure fairness, all methods are evaluated with  same multimodal input modalities (text, audio, and visual) whenever applicable.

Our proposed \textsc{SDT-VEGA} consistently outperforms all sota, achieving substantial improvements of \textbf{2.07\%--11.2\%} in accuracy and \textbf{1.5\%--9.89\%} in F1 score on IEMOCAP, and \textbf{2.02\%--8.3\%} in accuracy and \textbf{1.75\%--9.89\%} in F1 on MELD. 

\subsection{Baselines and Modality-Agnostic Evaluation}

To validate the modality-agnostic and architecture-agnostic nature of \textsc{VEGA} mechanism, we construct three unimodal baselines, each comprising a modality-specific feature extractor followed by a Transformer-based temporal modeling module. The three configurations correspond to visual, textual and acoustic modalities, utilizing DenseNet, RoBERTa and OpenSmile, respectively. For each baseline, we report performance on IEMOCAP under two settings: a standard version and an enhanced version with the \textsc{VEGA} branch.
As shown in Table~\ref{baseline}, integrating \textsc{VEGA} consistently improves performance across all modalities and most emotion categories. Specifically: (i) augmenting DenseNet (visual) with \textsc{VEGA} yields a \textbf{3.01\%} gain in F1 score and \textbf{1.61\%} in accuracy; (ii) incorporating \textsc{VEGA} into RoBERTa (text) leads to a \textbf{2.29\%} increase in F1 and \textbf{2.47\%} in accuracy; (iii) applying \textsc{VEGA} to OpenSmile (audio) results in a \textbf{2.54\%} improvement in F1 and \textbf{3.33\%} in accuracy. These results underscore \textsc{VEGA}'s general applicability and its effectiveness in enhancing affective representations across diverse modalities.

\subsection{Architecture-Agnostic Evaluation}

To assess the architecture-agnostic nature of \textsc{VEGA}, we integrate it into \textbf{MSRFG}, a graph-based dialogue model that stands in stark contrast to our transformer-based SDT backbone. As reported in Table~\ref{tab:msrfg-vega}, the VEGA-augmented variant, \textbf{MSRFG-VEGA}, consistently surpasses its baseline counterpart across both datasets. It yields absolute gains of \textbf{2.7\%} in accuracy and \textbf{1.9\%} in F1 on MELD, and \textbf{1.8\%} in accuracy and \textbf{1.6\%} in F1 on IEMOCAP. These results underscore VEGA’s strong generalizability and effectiveness when deployed on fundamentally different architectural paradigms.

\begin{table}[h]
\centering
\caption{Performance comparison between MSRFG and MSRFG-VEGA on MELD and IEMOCAP.}
\label{tab:msrfg-vega}
\setlength{\tabcolsep}{10pt}
\renewcommand{\arraystretch}{1}
{\fontsize{9pt}{9pt}\selectfont
\begin{tabular}{lcccc}
\toprule
\multirow{2}{*}{\textbf{Model}} & \multicolumn{2}{c}{\textbf{MELD}} & \multicolumn{2}{c}{\textbf{IEMOCAP}} \\
\cmidrule(lr){2-3} \cmidrule(lr){4-5}
 & Acc & F1 & Acc & F1 \\
\midrule
MSRFG         & 64.2  & 63.8  & 72.7  & 71.6 \\
\rowcolor{gray!20}   
\textbf{MSRFG-VEGA}    & \textbf{66.9}  & \textbf{65.7}  & \textbf{74.5}  & \textbf{73.2} \\
\bottomrule
\end{tabular}}
\end{table}

\subsection{Qualitative Analysis via t-SNE Visualization}

We visualize the learned representations using t-SNE to compare SDT and SDT-\textsc{VEGA}. The visualization reveals that SDT learns fragmented and overlapping emotion clusters, e.g., \textit{Neutral} is entangled with other classes, while \textit{Angry} and \textit{Frustrated} are difficult to distinguish. In contrast, SDT-\textsc{VEGA} forms compact, well-separated, and semantically coherent clusters for each emotion, with clearer boundaries and reduced ambiguity. These results suggest that \textsc{VEGA} encourages more discriminative, structured, and consistent representations, aligning with the observed performance gains.

\begin{figure}[htbp]
    \centering
    \includegraphics[width=1\linewidth]{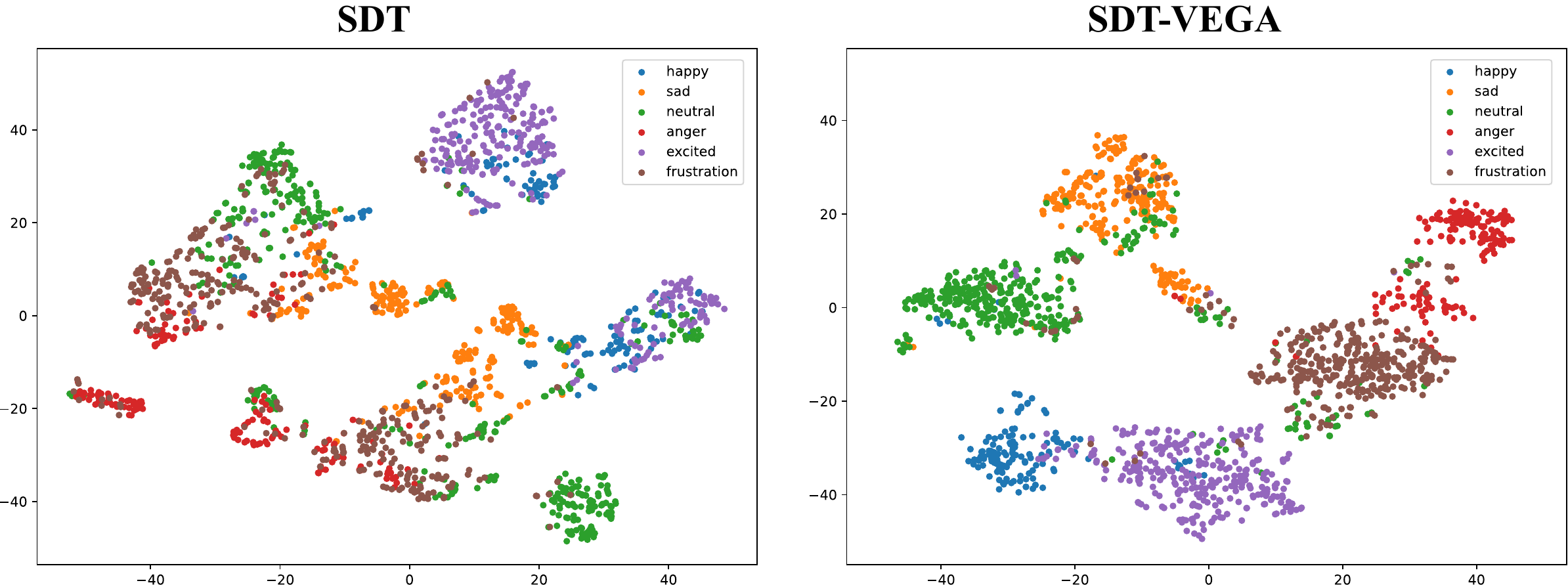}
    \caption{t-SNE of the fused feature on IEMOCAP test set.}
    \label{fig:tsne}
\end{figure}

\subsection{Computational Efficiency}

VEGA introduces \textbf{zero inference overhead}, as it is entirely removed during testing. During training, the additional computational cost is minimal: anchor features are precomputed once, and only a lightweight projection head with a cosine similarity loss is introduced. The base model, SDT, contains 79.7M parameters and requires 9.8 GFLOPs. Integrating \textsc{VEGA} adds only 4.7M parameters (a 5.9\% increase) and 0.2 GFLOPs (a 2\% increase). On a GeForce RTX 4090, the training time per epoch increases marginally from 26 seconds (SDT) to 27 seconds (SDT-\textsc{VEGA}), confirming the negligible overhead. Moreover, \textsc{VEGA} accelerates convergence:  best result is achieved at epoch 22 for \textsc{VEGA}, compared to epoch 39 for SDT.

\subsection{Ablation Study}

To investigate the contribution of individual architectural components in our proposed method, we conduct a series of ablation studies, summarized in Table~\ref{ablation}.

\textbf{Model Structure: }
Ablating \textit{positional embeddings} causes a clear performance drop, reflecting their role in capturing temporal structure. Removing \textit{speaker embeddings} leads to even greater degradation, underscoring the value of speaker-aware cues. Eliminating \textit{intra-modal transformers} moderately impairs performance, while removing \textit{inter-modal transformers} further reduces accuracy, confirming the importance of both modality-specific and cross-modal modeling. These results highlight the complementary contributions of temporal, speaker, and cross-modal cues to emotion recognition.

\textbf{Single vs. Dual Branch Design. }
We compare two fusion-level classification schemes: \textit{Single Branch}, where CLIP-space projection precedes classification and self-distillation, and \textit{Dual Branch}, which decouples Supervision and VEGA into separate heads. The dual-branch design yields a notable improvement of \textbf{1.63\%} in accuracy and \textbf{1.22\%} in F1. This gain highlights the benefit of separating optimization paths: Supervision Branch focuses on label discrimination, while VEGA Branch aligns features with external semantics.

\textbf{Impact of Anchoring and Distillation Losses}  
\textit{(1) Without multimodal anchoring ($\mathcal{L}_{\text{anc}}^{\text{fuse}}$).} Leads to a significant drop (72.61\% acc, 71.93\% F1), indicating that aligning the fused output with CLIP anchors is essential. Without it, training is dominated by classification losses, reducing semantic grounding.  
\textit{(2) Without modality-specific anchoring ($\mathcal{L}_{\text{anc}}^{(m)}$).} Removing this still yields strong results, likely because the anchor-based distillation implicitly guides unimodal features toward the semantically aligned fusion space.  
\textit{(3) Without CLIP-based distillation ($\mathcal{L}_{\text{CLIP-dist}}^{(m)}$).} Performance decreases despite retaining both uni- and multimodal anchor losses, confirming that anchor-based distribution matching is a crucial bridge between unimodal and fusion spaces.  
\textit{(4) Classification-only losses.} Using only classification and self-distillation ($\mathcal{L}_{\text{cls}} + \mathcal{L}_{\text{cls}}^{(m)} + \mathcal{L}_{\text{dist}}^{(m)}$) underperforms the full objective, underscoring the importance of semantic anchoring and CLIP-based visual priors.  
\textit{(5) Anchor-only setting.} Using only anchoring losses ($\mathcal{L}_{\text{anc}}^{\text{fuse}} + \mathcal{L}_{\text{anc}}^{(m)} + \mathcal{L}_{\text{anc-dist}}^{(m)}$) without any classification terms causes further deterioration, showing that while anchors provide structure, they cannot substitute for hard-label supervision.  
These results highlight that anchoring and classification play complementary roles. 
The full loss objective achieves the best performance by jointly enforcing semantic alignment and decision boundary learning.

\textbf{Impact of Teacher Selection}  
We investigate the effect of different teacher signals used in the two self-distillation objectives: anchoring-based distillation and classification-based distillation. While both aim to transfer fusion-level semantics to unimodal branches, they rely on distinct supervision signals. We conduct ablation studies by swapping their respective teacher:
\textit{(1) Anchoring-based distillation with classification teacher.}  
Replacing anchoring teacher with classification logits significantly reduces performance. 
\textit{(2) Classification-based distillation with anchoring-based teacher.}  
Conversely, using the anchoring fusion distribution to supervise the classification branch leads to further degradation. 

\textbf{Visual Sample Number in Anchor Construction}
We evaluate the impact of image diversity by varying the number of images used per class to construct anchors. Using only a single image per class yields the worst performance due to limited intra-class coverage, resulting in rigid and unrepresentative anchors. Increasing to 35 images brings a significant improvement by balancing diversity and consistency. However, using 100 or more images slightly degrades performance, likely due to over-diversification introducing noise and assignment instability. Excessively dispersed anchors reduce alignment precision and slow convergence under our similarity-thresholded matching strategy.

\textbf{Effect of Anchor Sampling Threshold $q$}
We examine how the anchor sampling threshold $q$ affects performance. Best results are achieved at $q = 0.2$, which balances exploration of diverse anchors with convergence toward stable assignments. Setting $q = 0$, which always selecting the center anchor, slightly reduces performance, likely due to limited intra-class variability modeling. Higher values (e.g., $q = 0.5$) lead to further degradation, as excessive randomness in anchor assignment hampers semantic consistency.

\textbf{Effect of CLIP Visual Encoder Choice}
We assess the impact of different CLIP visual encoders on the quality of emotion anchors. Among the tested backbones, ViT-L/14@336 achieves the best performance, outperforming smaller or lower-resolution variants such as ViT-B/16.
This improvement can be attributed to two factors: (1) \textit{larger model capacity} (ViT-L) enables better capture of high-level emotional semantics, and (2) \textit{higher input resolution} (336px) preserves subtle affective cues (e.g., facial micro-expressions) crucial for emotion grounding. In contrast, lower-capacity or lower-resolution models produce anchors with limited semantic granularity, weakening feature-to-anchor alignment.

\textbf{Effect of Modality and Modality Combinations}
We evaluate the contribution of individual modalities and pairwise combinations to overall performance. Among unimodal inputs, text achieves the highest performance, reflecting its strong correlation with emotion semantics. Audio performs moderately, while visual cues alone yield the lowest results.
When combining modalities, all pairs show consistent improvement. The best result is obtained from text + audio, indicating complementary strengths, text provides semantic precision, while audio adds prosodic and tonal cues.

\begin{table}[htbp]
\caption{Ablation study on various architectural and training configurations. All numbers are percentages.}
\label{ablation}
\centering
\resizebox{\columnwidth}{!}{
\begin{tabular}{lccc}
\toprule
\textbf{Ablation} & \textbf{Configuration} & \textbf{ACC} & \textbf{F1} \\
\midrule
\rowcolor{gray!20}   
\multirow{1}{*}{\textbf{\makecell[l]{Ours}}} 
& \textbf{VEGA} & \textbf{76.02} & \textbf{75.58} \\
\midrule
\multirow{4}{*}{\textbf{\makecell[l]{Model \\ Structure}}} 
& w/o positional embeddings & 74.63 & 74.60 \\
& w/o speaker embeddings & 73.78 & 73.94 \\
& w/o intra-modal transformers & 75.20 & 74.74 \\
& w/o inter-modal transformers & 74.81 & 74.80 \\
\midrule
\multirow{2}{*}{\textbf{\makecell[l]{Classification \\ Structure}}} 
& Single Branch & 74.39 & 74.36 \\
& \textbf{Dual Branch} & \textbf{76.02} & \textbf{75.58} \\
\midrule
\multirow{5}{*}{\textbf{Loss}} 
& \scriptsize w/o $\mathcal{L}_{\text{anc}}^{\text{fuse}}$ & 72.61 & 71.93 \\
& \scriptsize w/o $\mathcal{L}_{\text{anc}}^{(m)}$ & 74.81 & 74.19 \\
& \scriptsize w/o $\mathcal{L}_{\text{anc-dist}}^{(m)}$ & 74.64 & 74.17 \\
& \scriptsize $\mathcal{L}_{\text{cls}} + \mathcal{L}_{\text{cls}}^{(m)} + \mathcal{L}_{\text{dist}}^{(m)}$ & 73.41 & 73.38 \\
& \scriptsize $\mathcal{L}_{\text{anc}}^{\text{fuse}} + \mathcal{L}_{\text{anc}}^{(m)} + \mathcal{L}_{\text{anc-dist}}^{(m)}$ & 73.00 & 72.56 \\
\midrule
\multirow{2}{*}{\textbf{\makecell[l]{Teacher \\ (Self-distill.
)}}} 
& \small $\mathcal{L}_{\text{anc-dist}}^{(m)}$: $\text{Teacher} = \hat{\mathbf{y}} $ & 73.62 & 73.03 \\
& \small $\mathcal{L}_{\text{dist}}^{(m)}$: $\text{Teacher} = \hat{\mathbf{y}}_{\text{anc}}^{(m)} $ & 72.14 & 71.95 \\
\midrule
\midrule
\multirow{5}{*}{\textbf{\makecell[l]{Image Num. \\ (per class)}}} 
& 1 & 72.71 & 72.87 \\
& \textbf{35} & \textbf{76.02} & \textbf{75.58} \\
& 50 & 75.74 & 75.42 \\
& 100 & 74.93 & 75.34 \\
& 1000 & 75.11 & 74.98 \\

\midrule
\multirow{5}{*}{\textbf{\makecell[l]{Anchor \\ Sampling ($q$)}}} 
& $q = 0$ (Center) & 75.07 & 74.17 \\
& {\boldmath$ q = 0.2 $} & \textbf{76.02} & \textbf{75.58} \\
& $q = 0.5$ & 74.21 & 73.62 \\
& $q = 0.7$ & 73.22 & 72.80 \\
& $q = 1$ (Random) & 72.37 & 72.63 \\
\midrule
\multirow{4}{*}{\textbf{\makecell[l]{CLIP \\ Structure}}} 
& ViT-B/16 & 73.64 & 73.29 \\
& ViT-B/32 & 75.14 & 74.99 \\
& ViT-L/14 & 74.61 & 74.52 \\
& \textbf{ViT-L/14-336} & \textbf{76.02} & \textbf{75.58} \\
\midrule
\midrule
\multirow{6}{*}{\textbf{\makecell[l]{Modality}}} 
& Text & 68.52 & 68.63 \\
& Audio & 61.12 & 60.96 \\
& Visual & 44.92 & 45.04 \\
& Text + Audio & 73.34 & 72.98 \\
& Text + Visual & 70.57 & 70.13 \\
& Audio + Visual & 63.78 & 63.39 \\

\bottomrule
\end{tabular}}
\end{table}


\section{Conclusion}
In this work, we present a novel multimodal emotion recognition framework that integrates data-driven learning with semantically grounded supervision. At the core of our design is Visual Emotion-Guided Anchoring (VEGA), which aligns unimodal and fused representations with emotion semantic priors through a dual-branch architecture. The framework incorporates emotion-level anchors, a stochastic anchor sampling strategy, and modality-to-fusion self-distillation to enhance robustness and discriminability across modalities. Extensive experiments on the IEMOCAP and MELD datasets demonstrate consistent performance gains over strong baselines and state-of-the-art methods. Beyond accuracy, our approach offers a modular and psychologically inspired paradigm for multimodal emotion recognition, paving the way for future research in semantically structured and human-aligned affective computing.

\section*{Acknowledgement}
The work of Dimitrios Kollias has been supported by Amazon UK LTD under AIRNAW project.

\bibliographystyle{ACM-Reference-Format}
\balance
\bibliography{sample-base}

\end{document}